\documentclass[conference]{IEEEtran}
\IEEEoverridecommandlockouts

\usepackage{cite}
\usepackage{amsmath,amssymb,amsfonts}
\usepackage{algorithmic}
\usepackage{graphicx}
\usepackage{textcomp}
\usepackage{xcolor}
\usepackage{hyperref}

\def\BibTeX{{\rm B\kern-.05em{\sc i\kern-.025em b}\kern-.08em
    T\kern-.1667em\lower.7ex\hbox{E}\kern-.125emX}}

\begin{document}

\title{Latent Space Reinforcement Learning for Steering Angle Prediction}

\author{\IEEEauthorblockN{Qadeer Khan}
\IEEEauthorblockA{\textit{TUM and Artisense}}
\and
\IEEEauthorblockN{Torsten Sch\"on}
\IEEEauthorblockA{\textit{Audi Electronics Venture}}
\and
\IEEEauthorblockN{Patrick Wenzel}
\IEEEauthorblockA{\textit{TUM and Artisense}}
}

\maketitle

\begin{abstract}
Model-free reinforcement learning has recently been shown to successfully learn navigation policies from raw sensor data. In this work, we address the problem of learning driving policies for an autonomous agent in a high-fidelity simulator. Building upon recent research that applies deep reinforcement learning to navigation problems, we present a modular deep reinforcement learning approach to predict the steering angle of the car from raw images. The first module extracts a low-dimensional latent semantic representation of the image. The control module trained with reinforcement learning takes the latent vector as input to predict the correct steering angle. The experimental results have showed that our method is capable of learning to maneuver the car without any human control signals.
\end{abstract}

\section{Introduction}

Reinforcement learning (RL) is gaining interest as a promising avenue to training end-to-end autonomous driving policies. These algorithms have recently been shown to solve complex tasks such as navigation from raw vision-sensor modalities. However, training those algorithms require vast amounts of data and interactions with the environment to cover a wide variety of driving scenarios. The collection of such data if even possible is costly and time-consuming. Simulation engines can help to easily collect driving data at scale. By interacting with a simulator, reinforcement learning can be used to train models that can map vision inputs to steering commands. This idea has been applied in recent works on autonomous driving navigation~\cite{DosovitskiyCoRL2017,LiangECCV2018}. 

In this paper, we present an approach to learn driving policies based on a low-dimensional state space representation. The key idea is to abstract the perception system from the control model such that the control model can be trained and optimized independently~\cite{WenzelCoRL2018}. This is shown in Figure~\ref{fig:segcodercontrol}. The system is organized as follows. First, a perception model uses raw sensor readings captured by an RGB camera placed at the front of the car as inputs to the system. This model processes theses inputs and produces an output map containing a pixel-wise semantic representation of the scene. Second, the semantic map of the scene is fed to the control model to produce a low-dimensional vector. The advantage of using a semantic representation of the scene instead of raw camera images is described below:

\begin{figure}
  \centering
  \includegraphics[width=\linewidth]{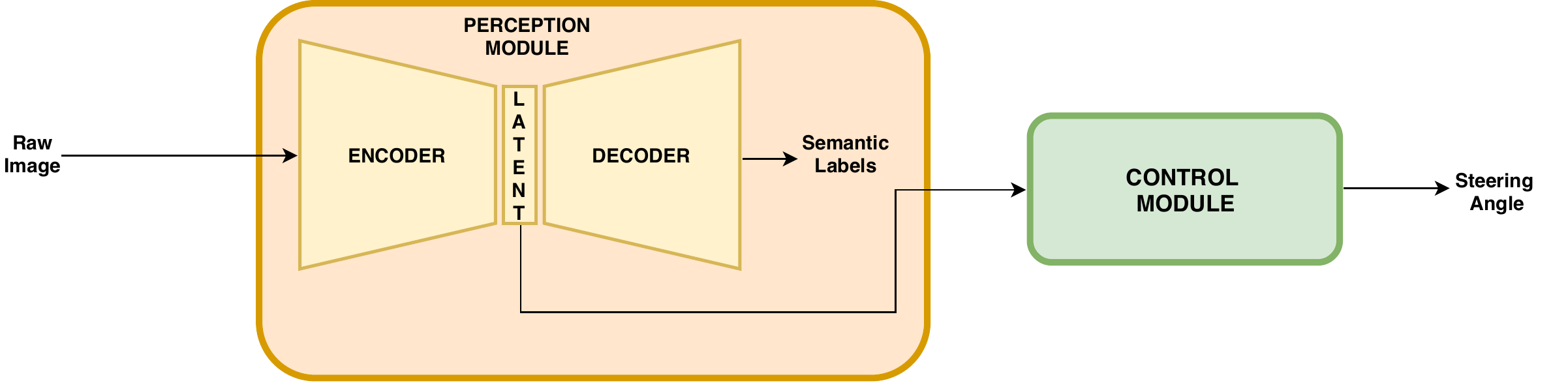}
  \caption{The perception model is trained as an encoder-decoder architecture without any skip connections. The encoder sub-module first embeds the raw image into a low-dimensional latent representation. The decoder sub-module reconstructs the semantic scene from this latent vector. We directly feed the semantic latent embedding as an input to the control module instead of the semantic labels.}
  \label{fig:segcodercontrol}
\end{figure}

\begin{figure}[ht]
  \centering
  \includegraphics[width=\linewidth]{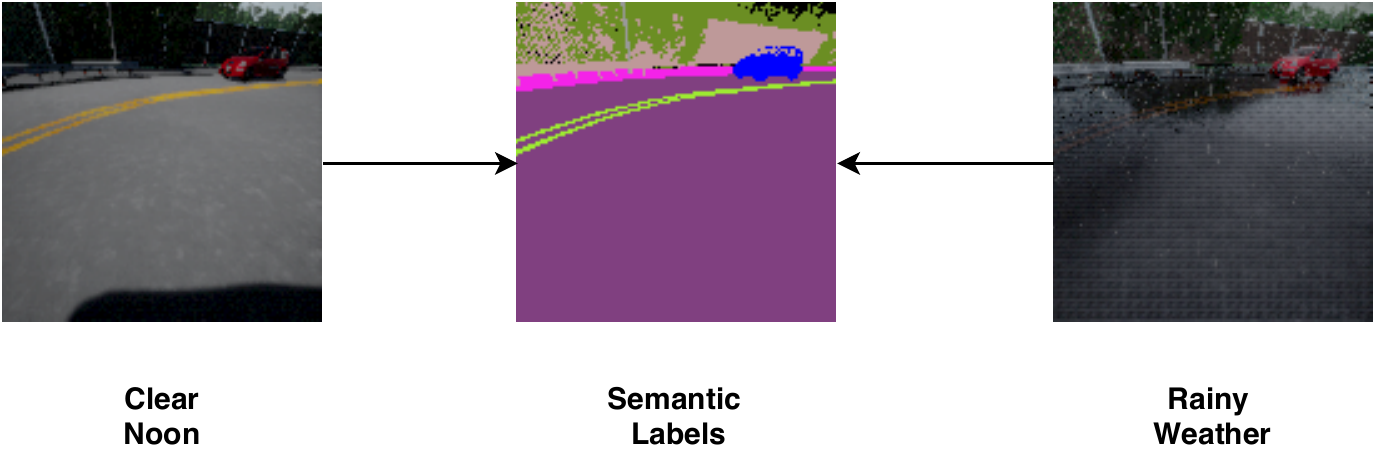}
  \caption{For the perception model we take in raw image data as obtained from the car's camera and output the semantic segmentation of the scene. Notice that irrespective of the weather condition the semantics of the scene remain the same. Since the perception model bears the burden of producing the correct semantic labels, the control model would be agnostic to changes in lighting, weather, and climate conditions.}
  \label{fig:whysegment}
\end{figure}

\begin{itemize}
\item Figure~\ref{fig:whysegment} shows how two weather conditions have different RGB inputs but the same semantic pixel labels. Hence, the control model does not separately need to learn to predict the correct steering commands for each and every weather condition.
\item The semantic labels can precisely localize the pixels of important road landmarks such as traffic lights and signs. The status/information contained on these can then be read off to take appropriate planning and control decisions.
\item A high proportion of the pixels have the same label as its neighbours. This redundancy can be utilized to reduce the dimensionality of the semantic scene. Hence, the number of parameters required to train the control model can then also be reduced. \end{itemize}

\begin{figure}
  \centering
  \includegraphics[width=\linewidth]{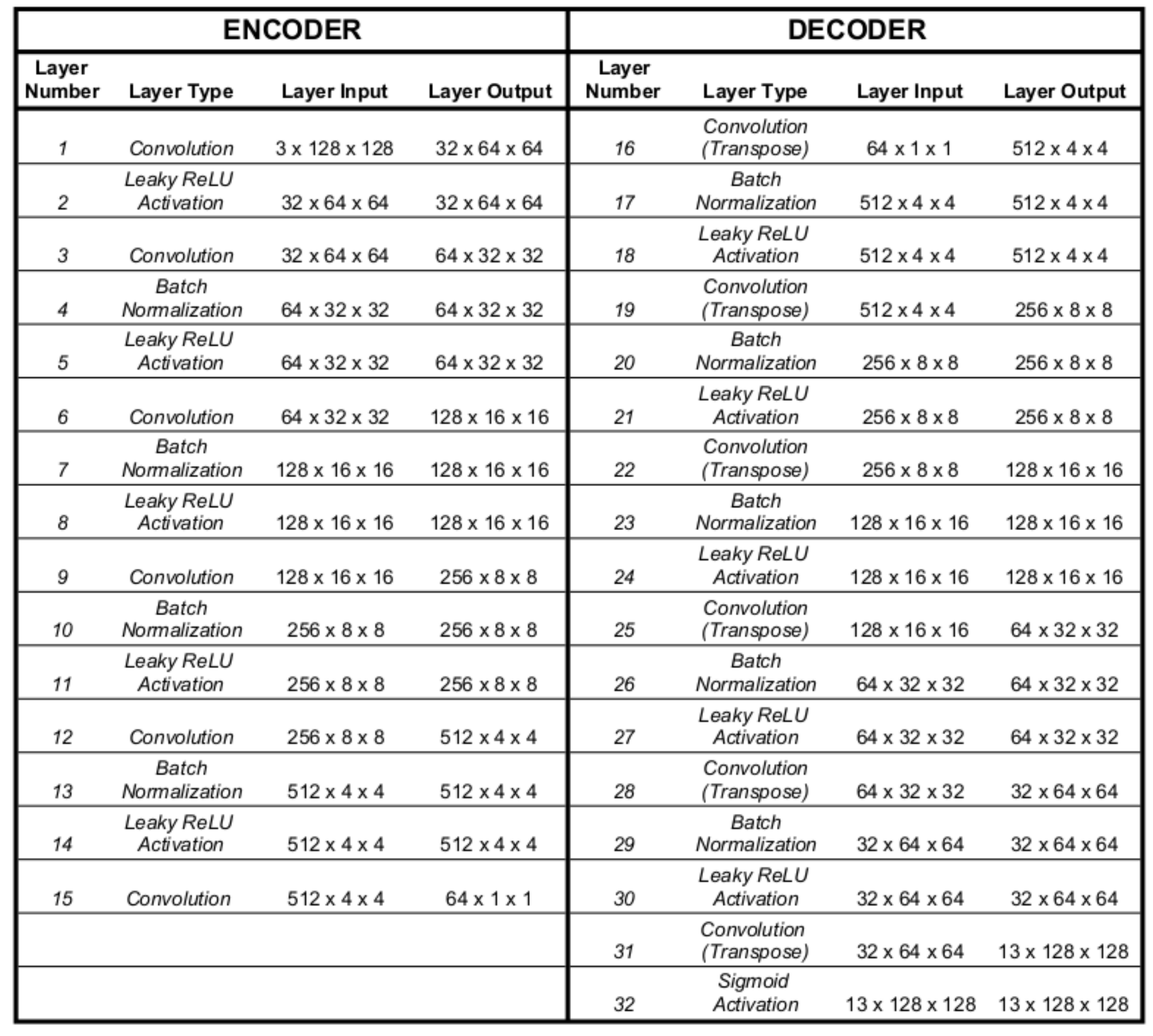}
  \caption{Encoder-decoder architecture used to train the segmentation perception model. The convolution layers numbered 15 and 16 have a kernel size of 4, stride of 1, and no padding. All other convolution layers have kernel size 4, stride of 2, and padding of 1. All the Leaky ReLU activation functions have a negative slope of $-0.2$. The output of the model has 13 channels with each corresponding to one of the semantic labels. The output of the last layer of the encoder (Layer 15) is fed to the control model to predict the correct steering direction.}
  \label{fig:segcoderArchi}
\end{figure}

\begin{figure}[ht]
  \centering
  \includegraphics[width=\linewidth]{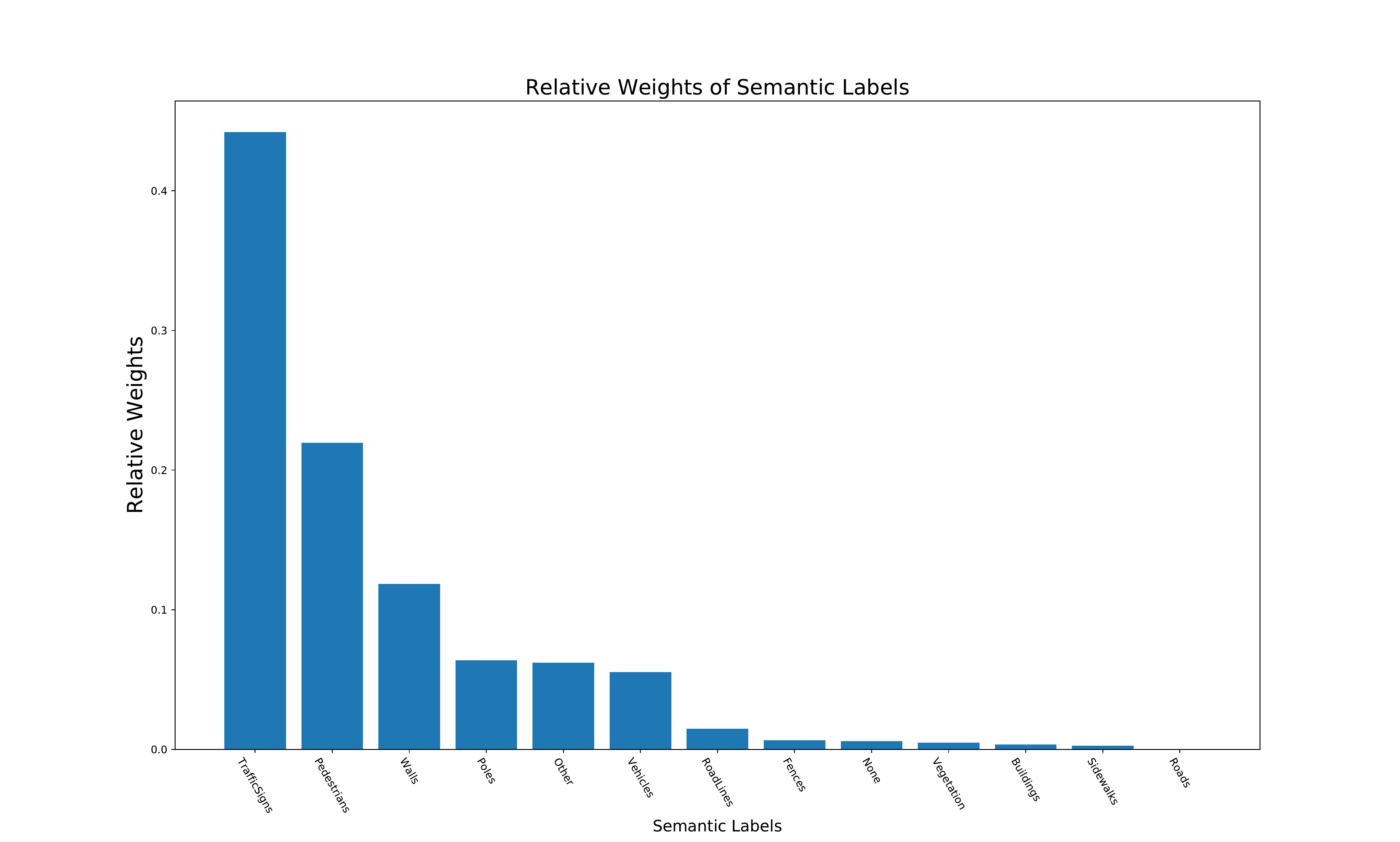}
  \caption{Relative weights of the semantic labels with their sum normalized to 1. The weights are inversely proportional to the frequency of occurrence of the corresponding label.}
  \label{fig:semanticlabels}
\end{figure}

The perception model, which is used to produce the semantic labels of the scene from the RGB camera is trained as an encoder-decoder architecture. The network architecture which is being used is a modified version of the one proposed by~\cite{LarsenICML2016}. The structure and the parameters of the model is described in Figure~\ref{fig:segcoderArchi}. The encoder first encodes the information contained in the input data to a lower dimensional latent vector. The decoder takes this latent vector and attempts to reconstruct the semantics of the scene. The output of the decoder is of the same size as the image, but having 13 channels with each representing the probability of occurrence of one of the semantic labels. Note that the semantic classes would be highly imbalanced since labels for commonly occurring entities such as the road would be more frequent than that for traffic lights. Therefore, the model is trained by minimizing the weighted sum of the categorical cross-entropy of each pixel in the image. The log-likelihood of a softmax distribution between predictions $p$ and targets $t$ is calculated as follows:

\begin{equation*}
    \mathcal{L}_i = \sum_j \, t_{i,j} \log(p_{i,j}) w_j,
\end{equation*}

where $i$ denotes the pixel and $j$ denotes the class. The weight $w_j$ of each semantic label is inversely proportional to its frequency of occurrence in the data set. Figure~\ref{fig:semanticlabels} shows the relative weights of the semantic labels with their sum normalized to 1.

\section{Related Work}

\noindent{\textbf{Semantic segmentation.}} The visual understanding of complex environments is an enabling factor for self-driving cars. The authors of~\cite{CordtsCVPR2016} provide a large-scale data set with semantic abstractions of real-world urban scenes focusing on autonomous driving. Semantic segmentation allows us to decompose the scene into a pixel-wise representation of classes relevant to interpret the world. This is especially helpful in the context of self-driving cars, \emph{e.g.} in order to discover drivable areas of the scene. It is therefore possible to segment a scene into different classes (\emph{e.g.} road and not road) and weight the different importance levels of distinct classes for driving systems~\cite{ChenIJCAI2017}. 

\noindent{\textbf{Imitation learning}.} The use of supervised learning methods to train driving policies for autonomous agents is a well-known and common approach. The first step towards using neural networks for the task of road following dates back to ALVINN~\cite{PomerleauNIPS1989} in 1989. In this work, a shallow neural network is used to map the input images and laser range findings directly to steering actions. More recently,~\cite{BojarskiArXiv2016} proposed to use an end-to-end deep convolutional neural network for the task of lane following. This approach demonstrated good results in relatively simple real-world driving scenarios. However, one major drawback of end-to-end learning system is the availability of enough labeled training data and therefore the possibility to generalize well to unseen scenes. 

\noindent{\textbf{Reinforcement learning.}} In reinforcement learning approaches, one crucial factor is the choice of the state space representation. A lot of prior work on deep reinforcement learning aim to learn purely from experience and discover the underlying structure of the problem automatically. This is a challenging problem, especially for sensorimotor control tasks as self-driving cars~\cite{YouBMVC2017,XuArXiv2018}. End-to-end vision-based autonomous driving models trained by reinforcement learning have a high computational cost ~\cite{DosovitskiyCoRL2017}. Training on a representative lower dimensional latent space allows for a less number of model parameters. The reduced number of parameters would allow for a significant speedup in training time. The authors of~\cite{LuckAAAI2016} proposed using variational inference to estimate policy
parameters, while simultaneously uncovering a low-dimensional latent space of controls. Similarly, the approach by~\cite{HaarnojaICML2018} has analyzed the utility of hierarchical representations for reuse in related tasks while learning latent space policies for reinforcement learning.

\section{Background}

We demonstrate our approach by training an autonomous agent on data obtained from a high-fidelity simulator with a reinforcement learning algorithm. We formulate the problem as a partially observable Markov decision process (POMDP). Below, we will cover the fundamentals of Q-learning.

In reinforcement learning, we assume an agent interacting with an environment. At each time step $t$, the agent executes an action $a_t \in \mathcal{A}$ from its current state $s_t \in \mathcal{S}$, according to its policy $\pi: \mathcal{S} \to \mathcal{A}$. The received reward at time $t$ which is obtained after interaction with the environment is denoted by $r_t: \mathcal{S} \times \mathcal{A} \to \mathbb{R}$ and transits to the next state $s_{t+1}$ according to the transition probabilities of the environment. The policy is considered optimal if it maximizes the expected sum of future rewards.

\section{Method}

\begin{figure}
  \centering
  \includegraphics[width=\linewidth]{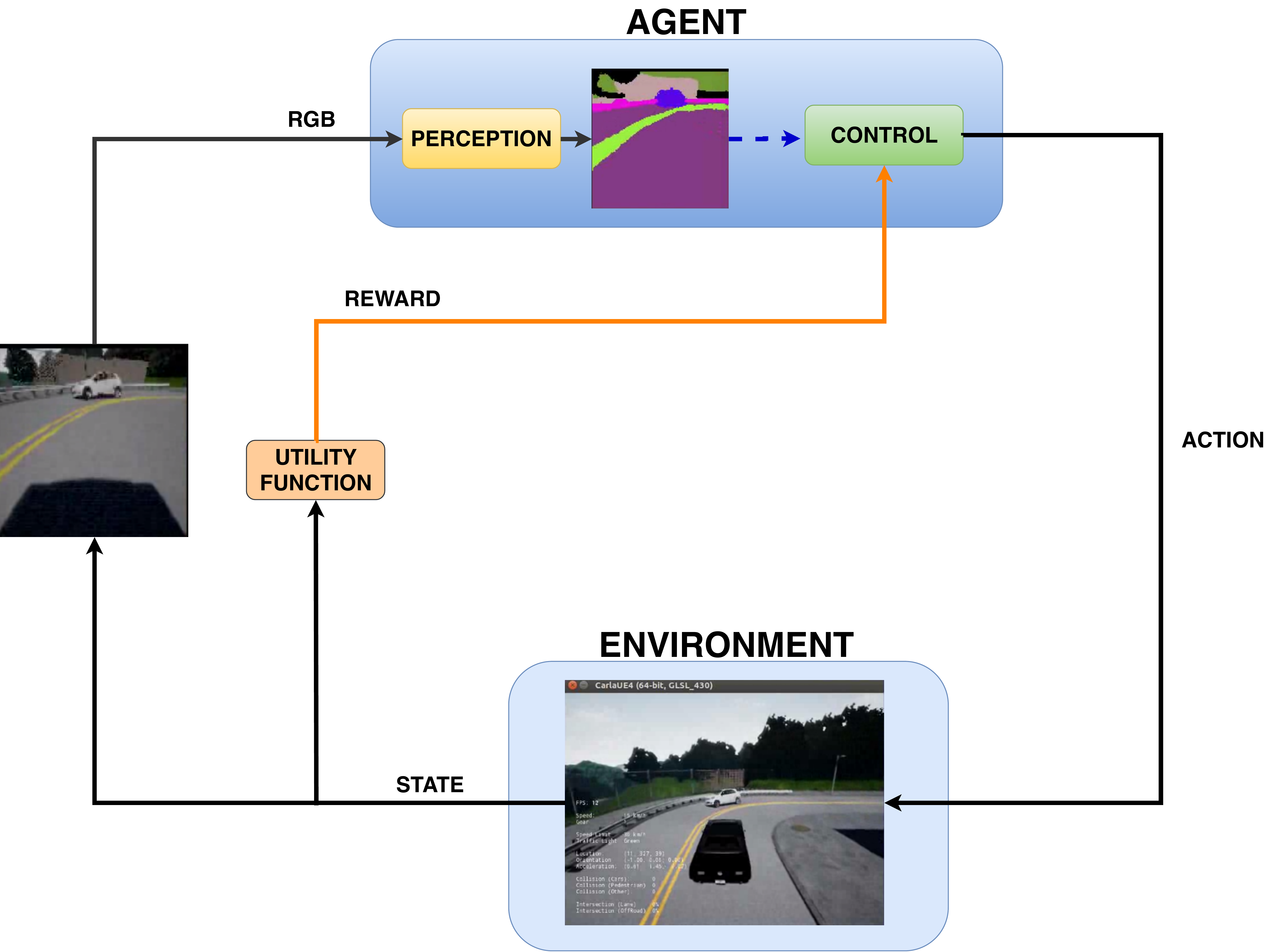}
  \caption{This figure shows the outline of training an agent using reinforcement learning. The CARLA~\cite{DosovitskiyCoRL2017} environment yields the state (RGB image) from the color camera placed at the front of the vehicle. The perception model of the agent converts the RGB image into the semantics of the scene. The dotted blue arrow shows that a low-dimensional representation of the semantic map is fed to the control model which decides on the appropriate action to be taken based on the current state. The agent interacts with the CARLA environment by executing the relevant action. This action changes the state of the car in the environment and hence a new state is generated and the cycle is repeated. Note, that there is no human supervision in the entire loop. The control model is therefore, trained from the reward signals that are dependent on the utility function formulated only once at the start of the training. The orange arrow to the control model indicates that its weights are updated based on these reward signals.}
  \label{fig:RLenv}
\end{figure}

Our method consists of a perception model that transforms images from the front-facing camera into a semantic representation of the scene. We are then able to train a reinforcement learning algorithm on the latent embedding of this state. In RL, instead of collecting supervised labels, we formulate a utility function. The car is then allowed to explore the environment at its discretion and accordingly learn from its experience. Figure~\ref{fig:RLenv} shows the steps involved in training an RL based agent.

The agent is provided with the state of the environment, on the basis of which it takes an action. Actions considered feasible are assigned a positive reward and adverse actions are assigned a negative reward. The rewards given are decided by the utility function defined once before the car starts exploration. With trial and error the car should learn how to maneuver itself for correct decision making without any explicit supervision. In our case, the CARLA simulator provides the RGB color image as the state. The perception model yields the semantics of the scene using this color image. The semantics are fed to the control model, which acts as the agent to decide the most appropriate action to be taken.

One limitation of RL is that it requires tremendous amount of experimental sessions to train~\cite{MnihICML2016}. In the paper by~\cite{DosovitskiyCoRL2017}, the RL algorithm on the CARLA simulator was trained for 12 days as opposed to 14 hours for imitation learning. Moreover, the input size of the image ($84 \times 84$) was also smaller in comparison with imitation learning ($200 \times 88$). Usually, a smaller input reduces the number of parameters and thus the number of sessions required for training~\cite{LuckAAAI2016}. As depicted in Figure~\ref{fig:segcodercontrol}, we feed the latent vector containing the semantic information to the control model instead of the complete scene. 

If we have the Markov property, then, irrespective of the past, the future state is only dependent on the current state. Hence, in our case, the next state $s_{t+1}$ determined by the action decision to be taken by the control model at time instance $t$ would only be dependent on the state $s_t$ (RGB image). This can mathematically be described as:

\begin{equation*}
P(s_{t+1} | s_{t}) = P(s_{t+1} | s_0, s_1, s_2, \dots, s_t)
\end{equation*}

In the following we describe the state space $s$, actions $a$ to be taken by the agent, reward $r$ received, discounting factor $\gamma$, as well as the transition probabilities $P$.

\begin{itemize}
\item \noindent{\textbf{State,} $s$} are the set of observations furnished by the environment on the basis of which the agent takes an action. As illustrated in Figure~\ref{fig:RLenv}, the CARLA simulator furnishes the color images from the monocular camera placed at the front of the car. The perception model converts this to a quasi-state latent semantic vector of size $64$ \emph{i.e.} $s \in \mathbb{R}^{64}$.

\item \noindent{\textbf{Action,} $a$} are the actions with which the agent can interact with the environment. The steering values are normalized between $-1$ and 1. To find the optimal policy we shall be using the off-policy based Q-learning algorithm which requires a discretized action space. We therefore, quantize the steering angle into 3 coarse values \emph{i.e.} $-0.4$, 0 and, $0.4$. 

\item \noindent{\textbf{Reward,} $r$} is the immediate reward or penalty received as a result of the agent executing an action and transitioning from current state $s_t$ to next state $s_{t+1}$. Since, we would like the car to drive for as long as possible, a reward of 1 is received at every step when the car is in the driving lane. If the car drives off-lane or off-road, a smaller reward or penalty is received. A reward of $-5$ is received either when the car is completely (100\%) in the other lane or (50\%) off-road. The episode ends if the car crashes with an obstacle, is of-road by more than 50\% or number of steps taken is greater than 500. The reward function at timestep $t$ can mathematically be expressed as:

\[
  r_t =
  \begin{cases}
                                   1, & \text{$r=0$ \& $l=0$,} \\
                                   1 + \alpha * l + \beta * r, & \text{if $r >0$ or $l > 0$,}
  \end{cases}
\]\label{eq:rewardFn}

where $r$, $l$ depict the percentage of the car off-road and off-lane, respectively. $\alpha = (R_r - 1) * l$ and $\beta = 4 * (R_r - 1) * r$, with $R_r = -5$. 

\item \noindent{\textbf{Discount factor,} $\gamma$} is a numerical value $\in [0, 1]$, which gives the relative importance of the immediate rewards in relation to future rewards it is expected to receive. A high value of $\gamma (= 1$) implies that immediate and future rewards are all equally important. This provides a rather farsighted view of the problem at hand. A low value of $\gamma (= 0$) gives a myopic view as it amplifies the significance of current rewards while all upcoming rewards can be discarded. Discounting is mathematically convenient for continuous tasks with non-terminal states as it keeps the expected sum of rewards bounded. Even in episodic tasks with terminal states, it still makes sense to have a discount factor, due to the uncertainty in future rewards associated with the stochastic nature of the environment. The discounted return, $G_t$ after execution of a sequence of state, action (and reward) pairs can be represented as:

\begin{equation*}
G_t = \sum_{k=0}^{N} \, \gamma^k  R_{t+k+1}
\end{equation*}

We have used $\gamma$ = 0.999 with $N = 500$ as the episode length. 

\item \noindent{\textbf{Transition probability,} $P$} furnishes the probability of the next state being $s_{t+1}$, given the current state $s_t$ and the action taken by the agent is $a_t$ \emph{i.e.} $P(s_{t+1} | s_{t})$. A transition probability tensor can be formed from each of the state, action and next state tuples. Each element in this tensor can be described by:

\begin{equation*}
P_{ss'}^a = P(s_{t+1} = s' | s_t = s, a_t = a)
\end{equation*}

These state transition probabilities, provide a model of the environment in the sense that based on the actions taken, the agent can anticipate the response of the environment. Algorithms utilizing such prior information about the environment are referred to as model-based. In our case the states are represented by a latent semantic vector in a continuous space and obtaining transition probabilities for such a large number of state-action-next state combination is not feasible even if possible. We therefore, resort to model-free methods, such as Q-learning. It is an indirect method of determining the optimal policy based on the value functions of the state action pairs.
\end{itemize}

\noindent{\textbf{Q-learning}}

We first define the value function for a given state $s$ as the expectation of the sum of future rewards when following a policy $\pi$.
 
\begin{align*}
    v_{\pi}(s) &= \mathbb{E} \left[G_t | s_t = s\right] \\
    &= \mathbb{E}\left[\sum_{k=0}^{N} \, \gamma^k  r_{t+k+1} | s_t  = s\right] \\
    &= \mathbb{E}\left[r_t + \sum_{k=1}^{N} \, \gamma^k  r_{t+k+1}|s_t  = s\right]. 
\end{align*}

The above equation can be recursively represented as a Bellman expectation equation:

\begin{equation}\label{eq:bellman_v}
    v_{\pi}(s) = \mathbb{E}\left[r_t + \gamma v_{\pi}(s_{t+1}) | s_t  = s\right].
\end{equation}

Loosely speaking the value of a state is the total sum of rewards the agent can receive if it starts from that state. Along similar lines, the Q-value for a particular state-action pair represents the expected sum of maximum future rewards. This is calculated by after having taken an action $a$ from state $s$, the agent follows the optimal policy there on-wards. Note that the action $a$ need not necessarily be the most optimal action. Analogous to Equation~\ref{eq:bellman_v}, the bellman optimality for the Q-values is defined by:
 
\begin{equation}\label{eq:bellman_q}
    Q(s, a)  = r_t + \gamma \arg\max_{a_{t+1}} Q(s_{t+1}, a_{t+1}).
\end{equation}
 
If the Q-values for each state-action pair are known, then at inference time we simply choose the action which yields highest Q-value for the given state. Note that whereas our action space is discrete, the state space is continuous. Therefore, it is not possible to tabulate the Q-values for each and every state-action pair. Rather, we use approximation methods which are one way of circumventing the scalability problem associated with Q-tables. We use neural networks to approximate the Q-value of an action associated with a certain state, hence the term Deep Q-Networks. This allows similar states which might not even have been seen during the training to be assigned action values close to those of similar observed states. Equation~\ref{loss_q} expresses the loss function used to update the weights of the control model. It is the square of the temporal difference (TD) between the predicted Q-values (red) and the target Q-values (blue). The target Q-value can themselves be determined by using the same control model and applying the Bellman equation.

\begin{equation}\label{loss_q}
    \mathcal{L} = \frac{1}{2} [\color{blue}(r_t + \gamma \arg\max_{a_{t+1}} Q(s_{t+1}, a_{t+1}) \color{black} - \color{red} Q(s,a)\color{black})]^2
\end{equation}
 
To expedite training and convergence of the deep Q-network, we use the enhancements proposed by~\cite{LillicrapICML2016}:

\begin{itemize}
 \item Separate target network
 \item Experience replay
\end{itemize}

\noindent{\textbf{Separate target network.}} Note that we are using the same network for finding both the predicted Q-value and the target Q-value. This results in a potentially dangerous feedback loop having a likelihood for creating instability. This problem can be addressed by using a separate network for calculating the target values which is independent from the primary Q-network. Thus, when the weights of the Q-network are updated, the target network remains fixed thus bringing relative stability. However, the target network also needs to be improved and this can be done after a certain number of episodes and independent from when the Q-network is updated. The target network has the same architecture as the primary network and is updated by simply copying the weights over from the primary network.
 
\noindent{\textbf{Experience replay.}} Note that the training samples we receive are on an episodic basis pertaining to one particular driving session. These samples would strongly be correlated and dependent on another and thus do not fulfill the requirement of being independent and identically distributed. Hence, if we train every episode independently, then the network is only learning about what the agent is currently doing in the environment for that episode. Therefore, when it sees a new state space, the error is high and so while updating the network it overfits to this current episode. The network’s output producing the Q-values would thus keep fluctuating between the episodes. This problem with training the network can be tackled by buffering the past experiences of the agent and randomly drawing a subset of samples from this buffer. The random sampling of past experiences allows the network to learn from a spectrum of different scenarios rather than just the current episode. This allows the network to generalize better.

\section{Experiments}

We evaluate our approach on the CARLA simulator. The perception module is trained using the encoder-decoder architecture by feeding the RGB images of size $128 \times 128 \times 3$ to the input of the encoder and reconstructing the $128 \times 128 \times 13$ semantic image from the output of the decoder. We collect data (RGB + semantic segmentation) with the car taking random driving decisions. This method of data collection with random exploration is representative of the state space that the agent will encounter while training with reinforcement learning.

The control module is trained with Q-learning. The vanilla Q-learning algorithm requires a discrete state and action space, wherein the combination of state action pairs are tabulated. We can do away with the requirement of the discrete state space by using the control module as a functional approximator. Hence, the continuous latent semantic vector produced by the output of the encoder can directly be fed to the control module as a representation of the state ($s \in \mathbb{R}^{64}$) of the environment.  An additional advantage of this is that every state does not necessarily have to be explored. Rather, states which might not even have been seen but are similar to those observed during training can be assigned similar Q-values, thus reducing the training effort.

We constrain the action space to only 3 possible action values. The advantage of this restriction is that we can speedup the training in comparison with training on a larger action space. Furthermore, we found that a combination of these 3 action values is enough to execute turns around corners. The architecture of the control module (also referred to as the deep Q-network) is described in Table~\ref{tab:controlArchi}. The output layer 11, which has 3 output neurons is predicting the Q-value for each of the 3 actions for a particular state represented by the latent semantic vector. Since the Q-values give the expected sum of future rewards, it would make sense to select the action corresponding to the highest Q-value, at inference time.  

\begin{table}[ht]
    \centering
    \caption{Architecture of the control model. Note that the input to the control module is a vector of size 64, corresponding to the size of the latent vector produced by the encoder of the perception module.}
    \resizebox{\linewidth}{!}{%
    \begin{tabular}{|cccc|}
        \hline
        \textbf{Layer Number} & \textbf{Layer Type} & \textbf{Layer Input} & \textbf{Layer Output} \\
        \hline
        1 & Fully connected & 64 & 100 \\
        \hline
        2 & ReLU activation & 100 & 100 \\
        \hline
        3 & Fully connected & 100 & 50 \\
        \hline
        4 & ReLU activation & 50 & 50 \\
        \hline
        5 & Fully connected & 50 & 25 \\
        \hline
        6 & ReLU activation & 25 & 25 \\
        \hline
        7 & Fully connected & 25 & 15 \\
        \hline
        8 & ReLU activation & 15 & 15 \\
        \hline
        9 & Fully connected & 15 & 8 \\
        \hline
        10 & ReLU activation & 8 & 8 \\
        \hline
        11 & Fully connected & 8 & 3 \\
        \hline
    \end{tabular}
    }
    \label{tab:controlArchi}
\end{table}

\begin{figure}[ht]
  \centering
  \includegraphics[width=\linewidth]{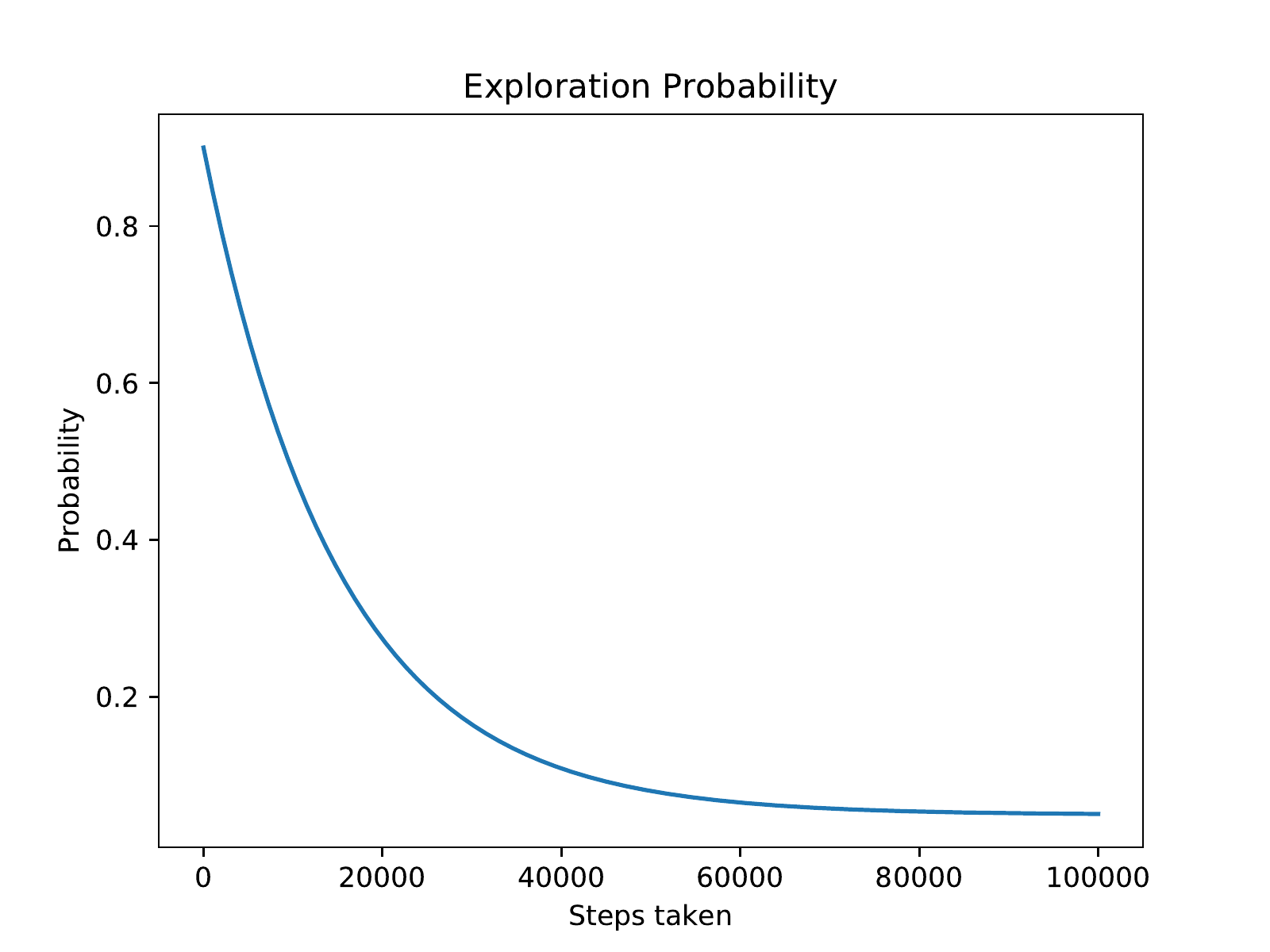}
  \caption{This figure shows the exploration probability as a function of the steps taken.}
  \label{fig:exploration}
\end{figure}

We use an experience buffer of size 7500 samples. All samples older that this number are discarded to make room for new (state, action, reward, and next state) tuples. Since the aim is to keep driving the car for as long as possible, a reward of 1 is given for every step the car is in the driving lane. Reward is reduced to $-5$ as a continuous function, if the car is 100\% in the other lane or 50\% off-road. The episode is terminated if the car crashes or has successfully executed 500 steps. The rewards are discounted by a factor of 0.999 at every step. The primary network is trained with a batch size of 512 at every step and the target network is updated every 256 steps. The simulation starts off with high exploration and gradually reduces towards exploitation. In the exploration phase, the actions are randomly selected whereas in exploitation, the action corresponding to the highest Q-value is chosen. The exploration starts off with a probability of $0.9$ and gradually reduces to less than $0.05$ after $10^5$ steps as shown in Figure~\ref{fig:exploration}. Table~\ref{tab:rl_parameters} enumerates values of some of the important parameters for training the Q-learning algorithm.
  
\begin{table}[ht]
    \centering
    \caption{This table summarizes information of some of the important parameters used for training the Q-learning algorithm.}  
    \begin{tabular}{|c|c|}
        \hline
        \textbf{Parameter} & \textbf{Value} \\
        \hline
        Size of state space $s$ & 64    \\
        \hline
        Size of action space $a$ & 3    \\
        \hline
        Size of experience replay buffer & 7500    \\
        \hline
        Discount factor $\gamma$ & 0.999    \\
        \hline
        Steps before updating target network & 256     \\
        \hline
        Batch Size & 512 \\
        \hline
        Maximum episode length $N$  & 500    \\
        \hline
    \end{tabular}
    \label{tab:rl_parameters}
\end{table}
    
\subsection{Discussion}
In order to visually check the performance of the trained policies, in the following some links to videos and their discussion is given. The video\footnote{\url{https://youtu.be/Sg3YkQEuE_k}} is an example of the car learning to execute a right turn with reinforcement learning. It can be observed that the steering of the car has many jerks. This is because that the action space only has 3 actions and the car's control is having to jump between the different steering values. These jerks could be avoided by using a larger action space at the expense of a longer exploration time. It can also be noted that, after having made the turn the car is in the other lane  most of the time and not in its driving lane. This is due to the fact that most of the exploration time was spent for making the turn. This can be improved by making the exploration probability also a function of the frequency of the states visited. Therefore, states visited less should have high exploration. A related point to note are the Q-values which sometime jump drastically between 2 similar states. These jumps are due to these states not having been explored enough.

Experiments also revealed that the reward function is formulated to be more well suited for right than left turns having barriers. Note that in case of 100\% exploration, the average steering value would be 0 and the car in expectation would move forward. In case of right turns, from the driving lane, the car eventually enters the other lane and then crashes into the barrier. All this while the reward function gradually reduces from 1 to $-5$ in a continuous manner.

\begin{figure}[ht]
  \centering
  \includegraphics[width=\linewidth]{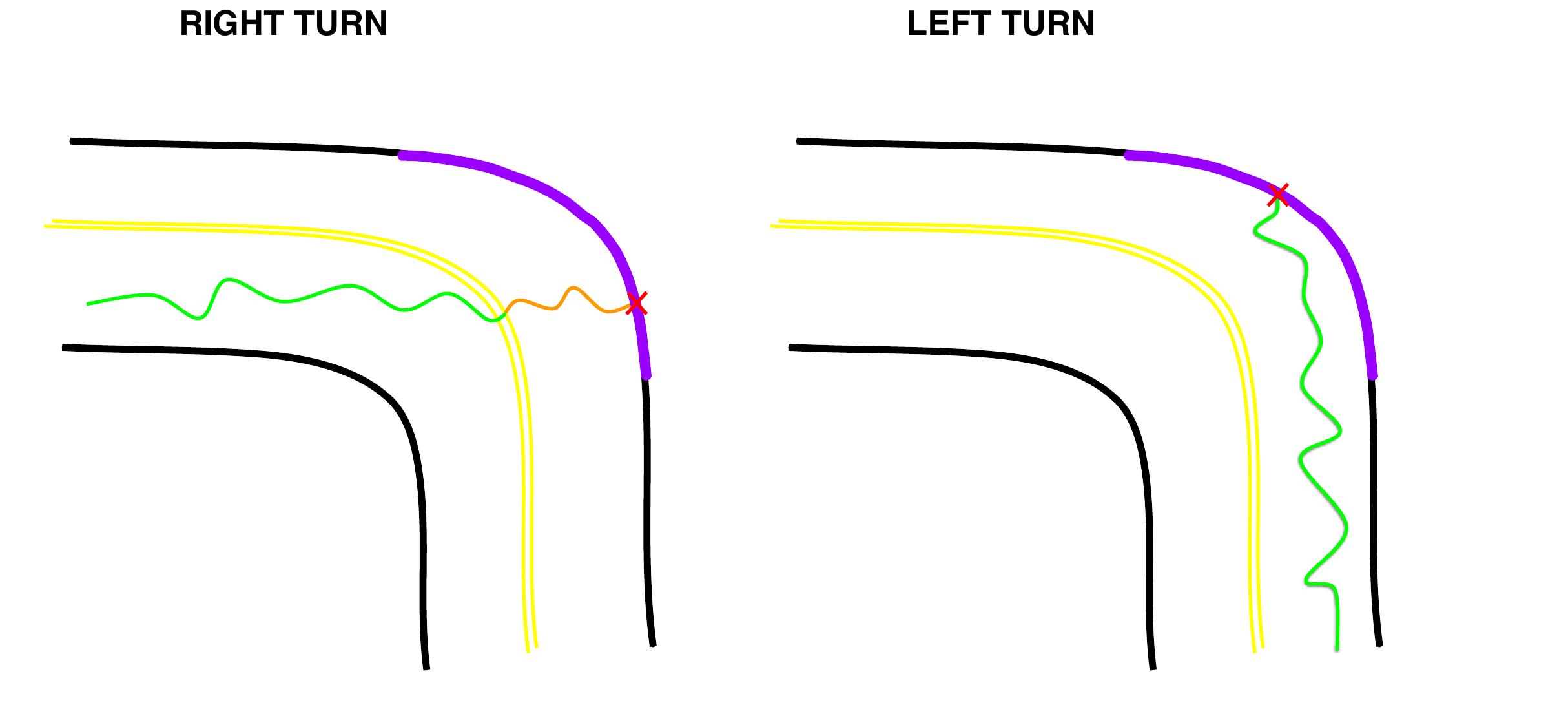}
  \caption{This figure shows the problem associated with executing a left turn using the current reward function. Green shows the trajectory of the car in the driving lane while receiving a reward of 1. Orange represents the trajectory of the car while in the opposite lane, wherein a reward between 1 and $-5$ is received as a continuous function. The exact reward is inverse relation to the percentage of the car in the other lane. The purple curve along the road edge, depicts the presence of a barrier/fence. The red crossed marking shows the point of impact of the car with the barrier in which scenario, a $-5$ reward is received. Note that in the case of a right turn the reward smoothly transitions as a continuous function from 1 to $-5$ at time of impact with the barrier. On the contrary for the left turn, there is a discontinuity in the reward where it suddenly transitions from 1 just one state before the impact to $-5$ at the time of impact. Such discontinuities in the reward function between similar states make the deep q-learning either slow to converge or even diverge.}
  \label{fig:rlLeftTurnProblem}
\end{figure}

This is shown in Figure \ref{fig:rlLeftTurnProblem} where the car receives a reward of 1 while following the green trajectory (driving lane). The reward smoothly transitions from +1 to $-5$ as it goes deeper into the other non-driving lane (orange trajectory). It eventually receives $-5$ reward when it collides with the barrier as shown by the red crossed marking. In the case of a left turn, the car in expectation would never move into the other lane, but suddenly crash into the barrier and there would be a discontinuity in the rewards which would jump from 1 to $-5$. Note that the deep-Q network is used as a function approximator, \emph{i.e.} similar states are assigned similar Q-values. The difference in states between time of impact of the crash and just one step before the crash is  effectively minimal. This minimal difference should also be reflected in the rewards these states receive, since the deep Q-network is trained from these reward signals. This holds true for right turns but for left turns there is a sudden change in rewards from 1 to $-5$. Hence, the network being a function approximator would have difficulties in assigning the appropriate Q-values to these states and would thus be either slow to converge or even diverge altogether.

\section{Conclusion}

In this paper, we have presented a framework to address the challenging problem of vision-based autonomous driving using reinforcement learning on a low-dimensional latent space representation. This offers the possibility of training the control module without any labeled steering angles. We observed that the car learned to execute a turn (albeit with some limitations) just by simply formulating a utility function at the start of the exploration phase. Hence, the need for an expert driver is completely eliminated. Based on the observations we also discussed some methods for further improving the RL algorithm. For future work, we are interested in transferring the policies trained in simulation to the real world.

\bibliographystyle{IEEEtran}
\bibliography{main}

\end{document}